\documentclass[11pt]{article}
\usepackage[a4paper,margin=1in]{geometry}
\usepackage{amsmath,amssymb,amsthm,mathtools}
\usepackage{graphicx}
\usepackage{booktabs,array}
\usepackage[hidelinks]{hyperref}

\newcommand{\R}{\mathbb{R}}
\newcommand{\one}{\mathbf{1}}
\newcommand{\KL}{D_{\mathrm{KL}}}
\newcommand{\softmax}{\operatorname{softmax}}
\newcommand{\CE}{\mathrm{CE}}
\newcommand{\im}{\operatorname{im}}
\newcommand{\tr}{\operatorname{tr}}
\newcommand{\Hsp}{\mathcal{H}}
\DeclareMathOperator*{\argmax}{arg\,max}
\DeclareMathOperator*{\argmin}{arg\,min}

\newcommand{\figbox}[3]{%
  \IfFileExists{#1}{\includegraphics[width=#2\textwidth]{#1}}%
  {\fbox{\parbox[c][#3][c]{#2\textwidth}{\centering\small [figure file \texttt{#1}]}}}}

\theoremstyle{plain}
\newtheorem{theorem}{Theorem}
\newtheorem{proposition}{Proposition}
\newtheorem{corollary}{Corollary}
\theoremstyle{remark}

\title{The Geometry of Multiclass Classification:\\
From Primal--Dual Probability Coordinates to Euclidean Realization}
\author{Mounir Beggas\\
\small Department of Computer Science, Faculty of Exact Sciences\\
\small LIAP Laboratory: Artificial Intelligence and Applications Laboratory\\
\small University of El Oued, Algeria}
\date{21 November 2026}

\begin{document}
\maketitle

\begin{abstract}
Multiclass classification associates each input with a categorical prediction and compares that prediction with a target label. Existing work on its geometry has concentrated on learned features, classifier weights, and the conditions under which a simplex equiangular tight frame (ETF) appears, notably in neural collapse. We study a complementary question: what geometry is already specified by the conventional one-hot, softmax, cross-entropy formulation, and how does it lead toward a Euclidean realization? We first develop probability distributions and relative logits as two coupled descriptions of the categorical model. Negative entropy and log-sum-exp are Legendre-dual convex potentials, and the cross-entropy objective is the KL divergence from a one-hot target to a prediction, equivalently a Fenchel--Young gap. To distinguish the geometric roles of direction and magnitude, we further analyze these coordinates through an angular--radial decomposition, after centering the logits to remove the common-shift redundancy of softmax. In centered logit coordinates, direction determines the pattern of relative log-odds and the class decision, while radius controls their magnitude and the concentration of the prediction. To investigate this geometry, we combine two complementary explanatory paths. The geometric path uses class-permutation symmetry to identify a barycenter-to-vertex probability segment whose inverse-softmax image is a canonical logit ray. The optimization path proves that cross-entropy uniquely selects this ray at every prescribed positive Euclidean logit radius. Their common origin in class symmetry yields a canonical equiangular configuration, which is recognized as a simplex ETF once Euclidean angles are measured. Finally, for a bias-free head, we characterize how logit directions and angular margins relate to feature geometry, and why balanced minimum-norm factorization, rather than logit geometry alone, produces aligned feature and classifier ETFs. Beyond recovering the known simplex configuration, the construction separates three levels of implication: the geometry specified by categorical supervision and class symmetry, the geometry conditionally selected by cross-entropy, and the geometry realized through Euclidean features and classifier weights. Making the relations and limits of implication between these levels explicit shows which properties of neural-collapse-type configurations are inherent to classification, which depend on the loss and a prescribed radius, and which require additional assumptions on the representation. The framework makes no claim about the convergence of neural-network training.
\end{abstract}


\section{Introduction}

In a $k$-class problem the label of an observation is a one-hot distribution, the classifier predicts another categorical distribution, and cross-entropy compares the two. A neural implementation, however, first produces real scores (logits) from learned Euclidean features and only then converts them into probabilities. Even before any question of training arises, this pipeline involves three different geometric objects: a simplex of distributions, a space of relative scores, and a feature space equipped with vectors and inner products.

\subsection{The geometric framework}\label{sec:framework}

We organize the investigation around three geometric levels, each with its own objects and its own notion of comparison: (i)~the \emph{probability geometry} of the categorical model, in which one-hot targets and predicted distributions lie on the simplex and are compared through the KL divergence; (ii)~the \emph{Legendre-dual logit geometry}, in which the same interior distributions are described by relative scores, or natural coordinates; and (iii)~the \emph{Euclidean realization geometry} of learned features and classifier weights, which generate those scores. This three-level organization is the framework we adopt to state the problem. It is not a theorem, and none of the individual spaces is new.

The first two levels are coupled through convex duality. A probability $p$ and a centered logit vector $z$ are two coordinate descriptions of the same interior categorical distribution, and the convex duality between negative Shannon entropy and log-sum-exp governs their relationship. A target label, which may lie on the boundary, is compared with the prediction through a single divergence that can be written from either side of this duality.

The third level is connected to the others in a different way. Features $h\in\R^d$ and classifier weights $W$ generate centered logits through $\Pi W h$, where $\Pi$ is the centering projector (introduced in Section~\ref{sec:primaldual}). This is a linear map chosen or learned by a model, not a second convex-dual coordinate change. It may have a kernel, distort angles, or fail to reach some logit directions. Our direction of explanation therefore runs from categorical supervision and its dual coordinates toward the additional conditions under which a corresponding Euclidean configuration can be realized. Understanding the geometry of classification thus requires examining the relations between the levels, and not only the objects within each.

The first two levels rest on established information-geometric foundations. The categorical simplex carries mixture coordinates, exponential-family natural coordinates, and the Fisher--Rao metric; negative Shannon entropy generates the KL divergence as a Bregman divergence, log-sum-exp is its convex conjugate, and softmax is the gradient map between them \cite{amari2000,nielsen2020}. Fenchel--Young losses make the resulting target--prediction gap explicit \cite{blondel2020}, and centered log-ratio coordinates from compositional data analysis likewise remove the unobservable common scale of a positive vector \cite{aitchison1982}. Aitchison geometry, Fisher--Rao geometry, and KL/Bregman divergence are nevertheless distinct metric or divergence structures on the same simplex, and our analysis does not identify them. We take these statements as established starting points and use them to build a single geometric interpretation, rather than treating KL, likelihood, and convex duality as competing explanations. The detailed construction is given in Section~\ref{sec:primaldual}.

\subsection{Related Work and Positioning}\label{sec:related}

Neural collapse is an important motivation for this work. Papyan, Han, and Donoho \cite{papyan2020} documented neural collapse (class-mean collapse, equiangular class means, aligned classifier weights, and the nearest-center decision rule) near the terminal phase of training, and gave a coding-theoretic rationale for the simplex structure.

A substantial theoretical literature has since derived simplex configurations from optimization models. Lu and Steinerberger \cite{lu2022} obtained simplex configurations for a constrained cross-entropy variational problem. Zhu et al.\ \cite{zhu2021} analyzed regularized unconstrained-feature models, Thrampoulidis et al.\ \cite{thrampoulidis2022} related label encodings to learned factors through a singular-value perspective, including imbalanced settings, and Hong and Ling \cite{hong2024} studied further imbalanced cross-entropy models. These results are derived for explicitly stated optimization models: the configurations obtained depend on the constraints or regularization imposed, on the model adopted for the features, and, in the imbalanced settings, on the class distribution. The relationship between labels and learned geometry has therefore been investigated extensively.

Angular-margin methods such as SphereFace \cite{liu2017} and ArcFace \cite{deng2019} deliberately modify the loss; we analyze directions and radial scales in ordinary softmax--CE without modification. Attraction--repulsion interpretations also appear in the neural-collapse literature \cite{xie2023}. Massion, Makhlouf, and Massart \cite{massion2026} study encoding and centering in MSE-based feature models; their results bear on the realization question but differ from the categorical cross-entropy construction developed here.

These studies establish important properties of learned representations, constrained optimization problems, and their simplex configurations. Our investigation takes a complementary direction: it constructs the geometric relationships specified by the classification formulation, beginning with targets and predictions, before examining how canonical configurations can be realized in Euclidean feature space. We therefore do not assume a simplex configuration in order to derive it again; we construct the relationships step by step and identify the simplex ETF when it appears in the argument. What we seek is an integrated, constructive geometric account, together with an explicit separation of what is implied at each level.

\subsection{Research Questions and Contributions}\label{sec:questions}

The shared notation can hide the differences. A class label is a vertex of the probability simplex, but regarding that vertex as a vector in $\R^k$ is an additional representation. A logit direction organizes relative class preferences, yet it is not a probability distribution. Two feature configurations can produce exactly the same logits while having different Euclidean angles. A coherent geometry of classification must therefore describe the objects in each space, the relations between the spaces, and the properties that those relations do or do not preserve.

This motivates the central question of the paper:
\begin{quote}
\emph{What geometry is specified by the mathematical formulation of multiclass classification, how is it expressed through probability and Legendre-dual logit coordinates, and under what conditions is this geometry realized in learned Euclidean representations?}
\end{quote}
A secondary question supplies an interpretive instrument:
\begin{quote}
\emph{How does angular--radial decomposition clarify the respective roles of class preference and probability concentration, and which directional relationships survive the passage from probability coordinates to logits and then to Euclidean features?}
\end{quote}

To investigate these questions within the coupled probability--logit description, we combine two complementary explanatory paths. The \emph{geometric encoding path} follows the class-symmetric segment from the uniform distribution to a label vertex and identifies its dual logit ray without reference to any loss. The \emph{optimization path} asks which direction on a fixed Euclidean logit sphere brings the prediction closest to the same vertex under KL divergence. Both express the same class-permutation symmetry of categorical supervision, so their agreement is structurally motivated; it is neither presented as a coincidence nor as a statement about learning trajectories.

The contribution is a geometric account supported by explicit identities, conditional theorems, and counterexamples. Convex duality, the centered label simplex, and balanced matrix factorization are established tools that we use as such.

First, we construct probability and centered-logit coordinates together and show that the KL/Bregman divergence and the Fenchel--Young gap are two faces of one classification objective. This separates what is inherent in the categorical model from what requires an additional metric choice. Second, we use angular--radial coordinates to distinguish class preference from log-odds magnitude, while keeping the Fisher metric distinct from the Euclidean norm on logits. Third, we show how class symmetry singles out a canonical ray and how cross-entropy uniquely selects the corresponding direction at every prescribed Euclidean logit radius; the equiangular configuration then appears as a consequence of the class orbit. Finally, we characterize how a bias-free Euclidean classifier can realize or distort these directions, including explicit non-implications between canonical logits and feature-space ETFs.

\begin{figure}[t]
\centering
\figbox{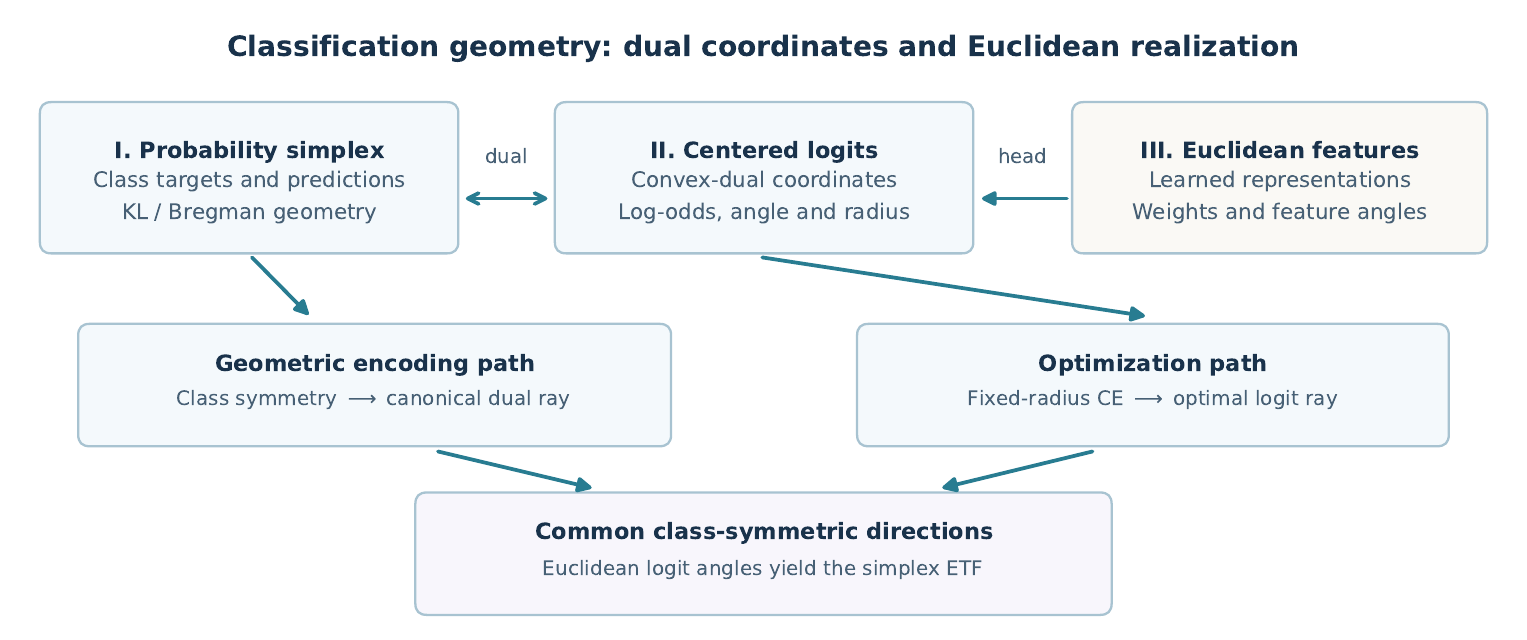}{0.95}{5cm}
\caption{The probability and centered-logit spaces are paired by an invertible dual-coordinate correspondence on the simplex interior. Learned features realize logits through a separate head that is generally not an isometry. The two paths characterize the canonical class rays; their Euclidean configuration and its realization are examined afterwards.}
\label{fig:overview}
\end{figure}

The paper is thus a constructive explanatory synthesis rather than a new general convergence theorem. Its aim is to state precisely which geometric relationships follow from classification itself, which follow from conditional selection by the loss, and which require further assumptions on representations. Figure~\ref{fig:overview} summarizes the architecture of the argument.

\subsection{Model, scope, and notation}

We study the conventional one-hot, softmax, cross-entropy model with $k\ge2$ classes. The final classifier is bias-free, $p(x)=\softmax(Wh(x))$ with $W\in\R^{k\times d}$. The conditional logit theorem treats a centered logit vector as a freely variable quantity; this is not equivalent to optimizing a constrained nonlinear network. The feature-realization analysis treats the final head as fixed when examining its geometry, and then studies factorizations of selected logit configurations.

\begin{center}
\begin{tabular}{@{}p{0.2\textwidth}p{0.72\textwidth}@{}}
\toprule
Symbol & Meaning\\
\midrule
$\Delta^{k-1},\ \Delta^\circ$ & Closed categorical probability simplex and its relative interior\\
$\xi$ & Unnormalized probability-mass vector, $\xi\in\R^k_{\ge0}\setminus\{0\}$, with $p_i=\xi_i/\sum_j\xi_j$\\
$e_c,\ b,\ u_c$ & One-hot target, uniform distribution $\one/k$, and $e_c-b$\\
$\Hsp,\ \Pi$ & Centered-logit subspace and orthogonal projector $I_k-\one\one^\top/k$\\
$F,\ A$ & Negative-entropy and log-sum-exp potentials\\
$v,\ p,\ z$ & Target distribution, prediction, centered natural logits\\
$\rho,\ q$ & Centered-logit Euclidean radius and unit direction\\
$r,\ s$ & Feature radius and unit direction, $h=rs$\\
$W$ & Classifier matrix with class rows $w_c^\top$\\
$M$ & Pullback quadratic form $W^\top\Pi W$\\
$H,\ Z$ & Matrices of class-representative feature rows and logits, $Z=HW^\top$\\
\bottomrule
\end{tabular}
\end{center}

The centering projector $\Pi=I_k-\tfrac1k\one\one^\top$ will appear in three geometric roles that do not have the same logical status: as the operation that measures deviations from the class barycenter, as the matrix underlying canonical logit configurations, and as the structure through which features and classifier weights may realize those configurations. Each role requires different assumptions, which are stated where the corresponding construction is developed (Sections~\ref{sec:primaldual}, \ref{sec:meeting}, and~\ref{sec:realization}); recognizing the same projector in all three places would conceal them.

\paragraph{Organization of the paper.}
Section~\ref{sec:primaldual} develops the coupled probability--logit geometry and its angular--radial interpretation. Sections~\ref{sec:encoding} and~\ref{sec:optpath} construct, respectively, the symmetry-preserving geometric path and the fixed-radius cross-entropy path. Section~\ref{sec:meeting} relates these paths and identifies the resulting simplex configuration. Section~\ref{sec:realization} then examines which logit structures a Euclidean classifier can attain and when its feature and weight geometry realizes them. Sections~\ref{sec:discussion} and~\ref{sec:conclusion} discuss the implications and conclude; supplementary arguments and examples are collected in the appendices.

\section{The Primal--Dual Geometry of Multiclass Classification}\label{sec:primaldual}

\subsection{Two coordinate spaces for one statistical prediction}

A label and a prediction are both categorical distributions, whereas a neural classifier produces a vector of real scores. To connect them geometrically we must determine what the probability and score descriptions share and what they do not. We begin with the spaces themselves, before giving them any Euclidean angles.

The closed probability simplex and its relative interior are
\begin{equation}
\Delta^{k-1}=\Big\{p\in\R^k:\ p_i\ge0,\ \sum_{i=1}^k p_i=1\Big\},\qquad
\Delta^\circ=\{p\in\Delta^{k-1}:\ p_i>0\}.
\label{eq:simplex}
\end{equation}
The one-hot labels $e_1,\dots,e_k$ are precisely its vertices, and finite softmax predictions lie in $\Delta^\circ$. Although the simplex has a standard affine embedding in $\R^k$, the information-theoretic comparison of two distributions does not require us to treat its drawn edges and angles as Euclidean distances.

Probability normalization has a useful property: a common positive scale carries no information. Every nonzero \emph{unnormalized probability-mass vector} $\xi\in\R^k_{\ge0}\setminus\{0\}$ (not a classifier weight) represents one distribution after normalization,
\begin{equation}
p_i=\frac{\xi_i}{\sum_j \xi_j},\qquad \xi\sim\lambda\xi\quad(\lambda>0).
\label{eq:normalization}
\end{equation}
The simplex is thus a section of the nonnegative cone, each positive ray giving one probability vector. These positive rays of unnormalized masses, which are exactly equivalent under normalization, should not be confused with logit rays, along which predicted probabilities generally change.

The standard learning objective is cross-entropy. For a target $v\in\Delta^{k-1}$ and an interior prediction $p$,
\begin{equation}
\CE(v,p)=-\sum_i v_i\log p_i=H(v)+\KL(v\,\|\,p),\qquad H(v)=-\sum_i v_i\log v_i .
\label{eq:ce}
\end{equation}
For a one-hot target this reduces to
\begin{equation}
\CE(e_c,p)=\KL(e_c\,\|\,p)=-\log p_c .
\label{eq:ce-onehot}
\end{equation}
This is also the categorical negative log-likelihood. The likelihood and divergence views are not competing theories; they are exact descriptions of the same expression. The KL form has the advantage of making the directed comparison between target vertex and prediction explicit. Over the closed simplex the divergence is minimized at $p=e_c$, so a finite softmax prediction can only approach that boundary vertex, whereas $\KL(p\|e_c)$ is infinite for every interior $p$. The direction of the divergence therefore matters.

The geometric question is now unavoidable: \emph{how can a comparison defined between probability distributions be expressed in the real-valued score coordinates of the network without changing the statistical objective?}

\subsection{Convex potentials linking probabilities and logits}

The softmax map is a modeling choice, not a consequence of the existence of one-hot labels, but the convex structure beneath it can be made exact. Consider negative Shannon entropy on the simplex,
\begin{equation}
F(p)=\sum_i p_i\log p_i,\qquad 0\log0:=0 .
\label{eq:F}
\end{equation}
A strictly convex potential assigns to each interior distribution a tangent hyperplane, and the gap between the potential and its tangent approximation at another point is a Bregman divergence. For interior $p$ and any target $v$,
\begin{equation}
D_F(v,p)=F(v)-F(p)-\langle\nabla F(p),v-p\rangle=\KL(v\,\|\,p).
\label{eq:bregman}
\end{equation}
The tangent is taken at the interior prediction, so the first argument may be the boundary vertex $e_c$. This gives the probability-side geometric meaning of cross-entropy without imposing Euclidean angles on the simplex.

Convex conjugacy supplies the corresponding potential in score coordinates:
\begin{equation}
A(z)=F^*(z)=\sup_{p\in\Delta^{k-1}}\{z^\top p-F(p)\}=\log\sum_i e^{z_i}.
\label{eq:A}
\end{equation}
The maximizer in \eqref{eq:A} is the softmax prediction, so
\begin{equation}
p=\nabla A(z)=\softmax(z),\qquad \nabla^2A(z)=\operatorname{diag}(p)-pp^\top .
\label{eq:softmax}
\end{equation}
Probability coordinates and natural logit coordinates are therefore linked by the gradients of a pair of convex potentials. In the interior, their tangent constructions carry the same statistical information, although the two coordinate systems do not share a Euclidean metric.

To make this correspondence one-to-one we must remove a redundancy. For a strictly positive $\xi$, write $z_i=\log\xi_i$ (all components must be positive for the logarithm to exist). This turns the multiplicative scaling freedom in \eqref{eq:normalization} into an additive common shift of the logits:
\begin{equation}
\xi\sim\lambda\xi\iff z\sim z+(\log\lambda)\one .
\label{eq:shift}
\end{equation}
Multiplicative rescaling of unnormalized masses and additive shifting of logits are thus the same redundancy, and neither changes the distribution. Radial scaling of a centered logit vector is different: it generally changes the prediction (Section~\ref{sec:angular}). To remove the redundancy we choose the zero-sum representative,
\begin{equation}
\Hsp=\{z\in\R^k:\ \one^\top z=0\},\qquad \Pi=I_k-\tfrac1k\one\one^\top .
\label{eq:H}
\end{equation}
On $\Hsp$, softmax is a diffeomorphism onto $\Delta^\circ$ with inverse
\begin{equation}
z=\Pi\log p .
\label{eq:clr}
\end{equation}
This inverse is the centered log-ratio representation of a positive composition. One-hot vertices have no finite coordinate under \eqref{eq:clr}; this is why the statistical problem has a boundary optimum while its centered-logit expression has no finite optimizer.

\subsection{The target-dependent objective as a single geometric gap}

Duality describes more than the relation between an input score and a probability vector; it also gives two equivalent descriptions of the objective. For an interior prediction $p=\nabla A(z)$, Fenchel equality reads $F(p)+A(z)=p^\top z$. Replacing $p$ by a prescribed target $v$, the supporting-hyperplane inequality gives $F(v)+A(z)-v^\top z\ge0$, and the gap is exactly
\begin{equation}
\KL(v\,\|\,\softmax(z))=A(z)+F(v)-v^\top z .
\label{eq:FY}
\end{equation}
Equations \eqref{eq:bregman} and \eqref{eq:FY} describe the same discrepancy, the first through the negative-entropy potential and the second through its conjugate. For $v=e_c$ we have $F(e_c)=0$, hence
\begin{equation}
L_c(z)=A(z)-z_c=\log\Big(1+\sum_{j\ne c}e^{z_j-z_c}\Big).
\label{eq:Lc}
\end{equation}
The target selects a linear tilt of the log-partition potential. A finite minimizer would require the tangent orientation $p=\nabla A(z)$ to equal $e_c$, which is impossible in the simplex interior; the gap instead tends to zero along suitable unbounded logit sequences.

For an interior target $v$, with $z_v=\Pi\log v$, the same relationship is the dual Bregman identity
\begin{equation}
\KL(v\,\|\,\softmax(z))=D_A(z,z_v).
\label{eq:dualBregman}
\end{equation}
A one-hot target has no finite $z_v$, so the Fenchel--Young form \eqref{eq:FY} remains the appropriate expression. These identities state what Legendre duality preserves, namely the statistical divergence and its convex structure. Euclidean distances and angles are not generally preserved.

The Fisher geometry sharpens this limit. At the uniform distribution $b=\one/k$, the Fisher--Rao metric on zero-sum probability tangent vectors $a,d$ is $g_b(a,d)=k\,a^\top d$, and at the corresponding logit $z=0$ the Fisher information is $\nabla^2A(0)=\Pi/k$. This local isotropy at the class-symmetric point does not identify the Fisher metric globally with a constant Euclidean metric. We therefore defer all statements about angles between class directions until those directions have been constructed.

\subsection{Relative-logit coordinates and angular--radial separation}\label{sec:angular}

With a unique centered coordinate $z$ for each interior prediction in hand, we can ask what the orientation and the magnitude of that coordinate describe. Legendre duality supplies the coordinate system; a Euclidean norm on $\Hsp$ then permits an angular--radial analysis. Take the standard Euclidean inner product on $\Hsp$ and write
\begin{equation}
z=\rho q,\qquad \rho=\|z\|_2\ge0,\quad q\in\Hsp,\ \|q\|_2=1\ \ (\rho>0).
\label{eq:polar}
\end{equation}
Then
\begin{equation}
\log\frac{p_i}{p_j}=z_i-z_j=\rho\,(q_i-q_j).
\label{eq:logodds}
\end{equation}
The unit direction $q$ specifies the pattern of relative log-odds, including the ordering of the classes, and the radius $\rho$ scales all differences simultaneously. A positive radial movement therefore stays within one decision cone:
\begin{equation}
\argmax_i\ \softmax(\rho q)_i=\argmax_i\ q_i,\qquad\rho>0 .
\label{eq:decision}
\end{equation}
This statement concerns the class decision, not the whole distribution. If $p=\softmax(z)$, then along the same logit ray
\begin{equation}
\softmax(tz)_i=\frac{p_i^{\,t}}{\sum_j p_j^{\,t}},\qquad t>0,
\label{eq:power}
\end{equation}
so the concentration of the distribution changes. More precisely, for $S(\rho)=H(\softmax(\rho q))$,
\begin{equation}
S'(\rho)=-\rho\,\mathrm{Var}_{\softmax(\rho q)}(q)<0\qquad(\rho>0).
\label{eq:entropyderiv}
\end{equation}
The radius thus increases concentration along any fixed nonzero direction. It does not necessarily increase confidence in the correct class: a direction pointing into a wrong decision cone becomes ever more confident in the wrong label. In this sense angular geometry organizes categorical decisions and radial magnitude controls their strength, while the full KL objective is sensitive to both.

The Euclidean metric is compatible with the symmetry of the labels. A constant positive-definite quadratic form on $\Hsp$ that is invariant under every class permutation must be a scalar multiple of the Euclidean form, because a symmetric matrix commuting with all coordinate permutations has the form $aI+b\one\one^\top$ and restricts to $aI$ on $\Hsp$. This is a uniqueness statement about constant permutation-invariant quadratic metrics; it does not replace the Fisher metric by the Euclidean one.

A one-hot target lies on the boundary, beyond the finite natural-coordinate chart. Every logit direction with a unique preferred class $c$ approaches $e_c$ as its radius grows, so the boundary objective does not by itself select a single ray. This observation motivates the two constructions that follow: which approach to $e_c$ is distinguished by symmetry alone, and which direction does the objective prefer at a finite, prescribed magnitude?

\subsection{The third level: a Euclidean realization, not another dual chart}

The standard bias-free model generates scores from features:
\begin{equation}
p(x)=\softmax(Wh(x)),\qquad W\in\R^{k\times d}.
\label{eq:head}
\end{equation}
Only $z=\Pi Wh$ affects the predicted probabilities. Unlike the nonlinear but invertible dual map between $\Hsp$ and $\Delta^\circ$, the linear head $\Pi W$ may have a kernel, may miss some logit directions, and generally does not preserve Euclidean angles. For a feature decomposition $h=rs$, positive rescaling preserves the predicted class because $z=r\,\Pi Ws$; this is one reason for using a bias-free head in the geometric analysis. Its exact angular and radial effects are treated in Section~\ref{sec:realization}, after the canonical class directions have been identified without assuming any feature configuration.

\section{Geometric Encoding Path: A Canonical Approach to Each Vertex}\label{sec:encoding}

\subsection{Many approaches, one class symmetry}

A one-hot target $e_c$ is a vertex of the closed simplex, while a softmax prediction remains in the interior. An interior point can approach the same vertex along infinitely many curves, corresponding in logit space to different rates at which the off-target probabilities vanish. Before any loss is minimized we can therefore ask: \emph{does the class encoding itself distinguish a particular approach to its target vertex?}

The $k$ class names play symmetric roles. Permuting the class coordinates exchanges the one-hot vertices, and the only distribution fixed by all such permutations is the uniform barycenter,
\begin{equation}
b=\tfrac1k\one .
\label{eq:barycenter}
\end{equation}
Once a target class $c$ is selected, the remaining $k-1$ classes are still interchangeable. Consider predictions that do not distinguish among the incorrect classes and give the target class at least as much probability as any of them. These distributions form exactly the barycenter-to-vertex segment
\begin{equation}
p_c(t)=(1-t)\,b+t\,e_c,\qquad 0\le t\le1 .
\label{eq:segment}
\end{equation}
This is a construction in affine probability geometry, requiring neither a divergence nor a Euclidean angle. The symmetry group fixing class $c$ determines the line, and the favored-class condition fixes its orientation; without that condition the same symmetry-fixed line continues in the opposite direction, to distributions in which class $c$ is less probable than its competitors.

The segment \eqref{eq:segment} is canonical, not exhaustive: it is the approach to $e_c$ that treats every competitor identically, and other approaches, including ones that favor some incorrect classes, are possible. The natural next step is to describe this distinguished segment in the dual coordinates of Section~\ref{sec:primaldual}.

\subsection{The dual image: a probability segment becomes a logit ray}

\begin{figure}[t]
\centering
\figbox{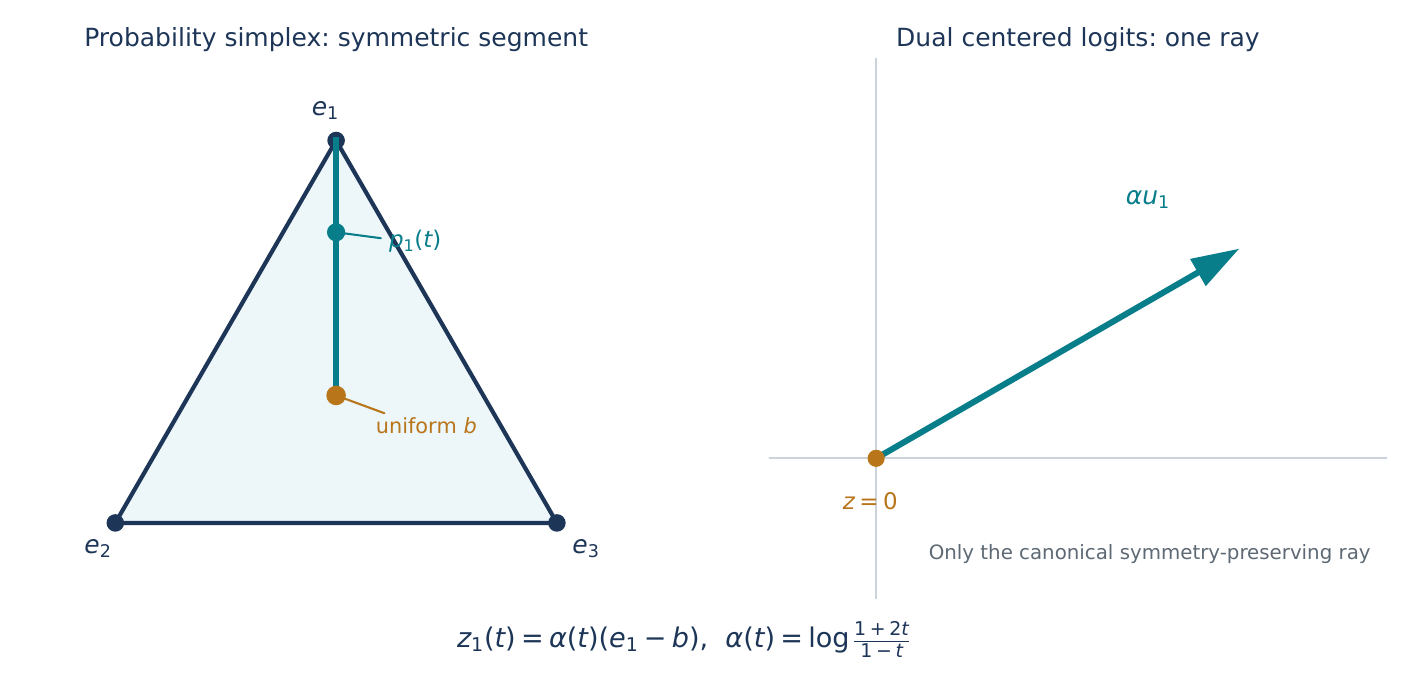}{0.9}{5cm}
\caption{Left: the label-preserving symmetry distinguishes the segment from the uniform barycenter to the favored vertex. Right: centered inverse softmax represents that segment as a straight positive logit ray, $z_1(t)=\alpha(t)(e_1-b)$ with $\alpha(t)=\log\frac{1+2t}{1-t}$ for $k=3$. The figure does not suggest that arbitrary lines in probability space map to lines, nor that the map preserves Euclidean angles.}
\label{fig:segment}
\end{figure}

Along \eqref{eq:segment} the target and competitor probabilities are
\begin{equation}
p_{c,c}(t)=\frac{1+(k-1)t}{k},\qquad p_{c,j}(t)=\frac{1-t}{k}\quad(j\ne c).
\label{eq:segcoords}
\end{equation}
All off-target probabilities coincide, so their mutual log-ratios vanish and every target-versus-competitor log-ratio takes the same value. Applying the centered inverse-softmax map \eqref{eq:clr} gives
\begin{equation}
\boxed{\,z_c(t)=\Pi\log p_c(t)=\alpha(t)\,u_c,\qquad u_c=e_c-b,\qquad \alpha(t)=\log\frac{1+(k-1)t}{1-t}\,.}
\label{eq:ray}
\end{equation}
For $0<t<1$ we have $\alpha(t)>0$, and $\alpha(t)\to\infty$ as $t\uparrow1$. The nonlinearity of the dual map reparametrizes $t$ as $\alpha(t)$, but the \emph{locus} becomes a single positive ray in centered logits. Class symmetry has thus selected the direction $u_c$ without any optimization criterion (Figure~\ref{fig:segment}).

The construction also shows that the boundary has no finite dual coordinate. As the prediction approaches the target along its symmetric segment, the natural coordinate runs to infinity while its direction stays fixed. The dual ray is therefore a useful object for understanding classification, without any assumption that an actual training trajectory follows it.

\subsection{What is canonical about the approach to the boundary?}

For $k=3$, the logit rays $t(2,-1,-1)$ and $t(2,0,-2)$ both lead under softmax to $e_1$ as $t\to\infty$. The first keeps the two incorrect classes equivalent; the second does not. More generally, every fixed centered $q$ with $q_c>q_j$ for all $j\ne c$ leads to the same vertex as its radius grows.

This observation is essential for the logic of the paper. Label geometry does not compel every vanishing-loss sequence to have a fixed orientation; it distinguishes \emph{one permutation-symmetric direction within an entire cone of successful directions}. The optimization path asks whether the ordinary classification objective assigns this distinguished ray a special status at finite magnitude.

\section[Optimization Path: The Geometric Preference of Cross-Entropy]{Optimization Path: The Geometric Preference of\\ Cross-Entropy}\label{sec:optpath}

\subsection{From target--prediction discrepancy to a centered force field}

The classification objective already interpreted as a Bregman tangent gap also has a differential description in logit coordinates. For a fixed target $v$ and $p=\softmax(z)$,
\begin{equation}
\nabla_zL(z;v)=p-v,\qquad \nabla_z^2L(z;v)=G(p)=\operatorname{diag}(p)-pp^\top .
\label{eq:grad}
\end{equation}
Since target and prediction are normalized distributions,
\begin{equation}
\one^\top(p-v)=0,
\label{eq:centeredgrad}
\end{equation}
so the loss gradient lies in the centered hyperplane automatically and has no component along the unobservable common shift. For a one-hot label the negative gradient decomposes as
\begin{equation}
-\nabla_zL_c=e_c-p=(1-p_c)\,e_c-\sum_{j\ne c}p_j\,e_j .
\label{eq:attrep}
\end{equation}
The positive target component balances the total negative competitor mass. This is a precise attraction--repulsion decomposition in logit coordinates, not a literal physical force and not a statement about individual feature trajectories. If the logits are realized as $z=Wh$, the corresponding feature gradient is
\begin{equation}
\nabla_hL_c=W^\top(p-e_c)=\sum_jp_jw_j-w_c,\qquad -\nabla_hL_c=w_c-\sum_jp_jw_j .
\label{eq:featgrad}
\end{equation}
The classifier weights determine which feature-space directions result from the same residual, and a nonlinear network adds further Jacobians and coupling between samples. For this reason the optimization path developed here concerns the geometric objective, not a universal description of training dynamics.

\subsection{Why the metric must be stated}

The Hessian $G(p)$ in \eqref{eq:grad} also represents Fisher information in natural coordinates. In probability coordinates, the Fisher--Rao metric on zero-sum tangent vectors $a,d$ is
\begin{equation}
g_p(a,d)=\sum_i\frac{a_id_i}{p_i}.
\label{eq:fisher}
\end{equation}
These are corresponding forms of an information-geometric metric; they are not the constant Euclidean metric used below to define the spherical constraint. The distinction is already visible in formal gradient flows. Euclidean logit descent induces
\begin{equation}
\dot z=-(p-v),\qquad \dot p=-G(p)(p-v),
\label{eq:flow1}
\end{equation}
whereas Fisher natural-gradient descent on the simplex yields $\dot p_{\mathrm{FR}}=v-p$. The comparison only demonstrates that the metric matters; it makes no claim about temporal training regimes or the convergence of neural networks.

\subsection{Conditioning on a radius as a geometric question}

A logit direction in the correct-class decision cone satisfies $q_c>q_j$ for every $j\ne c$, and every such ray produces predictions approaching $e_c$ as its radius tends to infinity. The free-logit cross-entropy objective therefore has many successful asymptotic directions, and sending the radius to infinity cannot distinguish a canonical one. The question becomes informative when directions are compared at equal Euclidean magnitude:
\begin{quote}
\emph{Among centered logit vectors of a prescribed positive radius, which direction brings the predicted distribution closest to the target under the original KL divergence?}
\end{quote}
This is neither a hidden regularization nor an assertion that training keeps the logit norm constant. It is a controlled geometric comparison: a sphere isolates orientation so that the loss can be studied as a function of direction alone, and the resulting direction can be compared with the canonical ray obtained without any loss.

\subsection{Exact angular selection at fixed logit radius}

Let $S_\rho=\{z\in\Hsp:\ \|z\|_2=\rho\}$ for $\rho>0$.

\begin{theorem}[Unique fixed-radius CE direction]\label{thm:direction}
For $k\ge2$ and every class $c$, the problem
\begin{equation}
\min_{z\in S_\rho}L_c(z)=\min_{z\in S_\rho}\big(A(z)-z_c\big)
\label{eq:thm1}
\end{equation}
has the unique solution
\begin{equation}
z_c^\star(\rho)=\rho\sqrt{\frac{k}{k-1}}\Big(e_c-\frac1k\one\Big).
\label{eq:zstar}
\end{equation}
Its normalized direction is independent of the prescribed radius,
\begin{equation}
q_c^\star=\frac{z_c^\star(\rho)}{\rho}=\sqrt{\frac{k}{k-1}}\,u_c,\qquad u_c=e_c-b,
\label{eq:qstar}
\end{equation}
and the minimum loss at radius $\rho$ is
\begin{equation}
\ell^\star(\rho)=\log\Big(1+(k-1)\exp\Big[-\sqrt{\tfrac{k}{k-1}}\,\rho\Big]\Big).
\label{eq:ellstar}
\end{equation}
\end{theorem}

\begin{proof}
The centering constraint gives $\sum_{j\ne c}z_j=-z_c$. For a fixed target coordinate, the competitors' exponential contribution is smallest when their logits agree; Jensen's inequality formalizes this:
\begin{equation}
\sum_{j\ne c}e^{z_j-z_c}\ \ge\ (k-1)\exp\Big(-\frac{k}{k-1}\,z_c\Big).
\label{eq:jensen}
\end{equation}
Next, the target coordinate on the Euclidean sphere is bounded by its projection on the centered target direction. Since $z_c=\langle z,u_c\rangle$, Cauchy--Schwarz gives
\begin{equation}
z_c\le\rho\|u_c\|_2=\rho\sqrt{\frac{k-1}{k}} .
\label{eq:cs}
\end{equation}
The right-hand side of \eqref{eq:jensen} is decreasing in $z_c$, so substituting \eqref{eq:jensen} and \eqref{eq:cs} into the dual loss \eqref{eq:Lc} yields the lower bound \eqref{eq:ellstar}. Equality requires both equal competitor logits and positive alignment $z\parallel u_c$; together with $\|z\|=\rho$ this gives exactly \eqref{eq:zstar}, which establishes existence and uniqueness.
\end{proof}

The proof exposes the geometric mechanism rather than merely producing a minimizer. Symmetry indicates which direction could be canonical; Jensen's inequality identifies the optimal equality among competitors; Cauchy--Schwarz uses the Euclidean radius to maximize the target advantage; and the two equality conditions pick out the same ray. Convexity of the loss alone would not suffice to infer uniqueness on the nonconvex sphere.

It is equally important to state what the theorem does not assert. It does not follow from Legendre duality alone, and it does not claim that ordinary gradient descent follows $q_c^\star$. Its content is the exact conditional optimality of a class-symmetric logit direction, with its orientation, uniqueness, and radial loss profile.

\section{The Meeting of the Paths and the Resulting Simplex Geometry}\label{sec:meeting}

\subsection{One class symmetry, expressed in two ways}

The loss-independent construction of Section~\ref{sec:encoding} produced $z_c(t)=\alpha(t)u_c$, while Theorem~\ref{thm:direction} selected $z_c^\star(\rho)=\rho\,u_c/\|u_c\|$. Both paths therefore distinguish the same positive ray in the canonical centered-logit coordinates. This agreement is not unexplained: both problems respect the permutations that fix the target class and exchange its competitors.

To make the role of symmetry precise, note that the subgroup $S_{k-1}$ of these permutations fixes, within $\Hsp$, exactly the line $\operatorname{span}\{u_c\}$. An invariant vector must have a common off-target coordinate, and the zero-sum condition leaves one free scalar. Consequently, if an invariant optimization problem on a centered logit sphere has a unique minimizer, that minimizer lies at one of the two intersections of this line with the sphere. Symmetry alone does not guarantee existence, uniqueness, or the correct sign; for cross-entropy these are established by Theorem~\ref{thm:direction}.

The geometric path thus shows how the canonical line is carried from a symmetry-preserving probability segment into dual coordinates, and the optimization path shows which orientation on that line best realizes the target at a given magnitude. Together they exhibit one underlying geometric organization in two descriptions of classification.

\subsection{From the orbit of class directions to an equiangular configuration}

Only now that the canonical directions have been constructed do we introduce the Euclidean angles between them. Relabeling the classes permutes coordinates, so the $k$ unit directions $q_c^\star$ form a single orbit of the class-permutation group. All have equal length, and since the symmetry carries any ordered pair of distinct labels to any other, all pairwise inner products are equal; their sum vanishes because each direction is centered. The standard coordinate calculation confirms this:
\begin{equation}
\sum_{c=1}^ku_c=0,\qquad\|u_c\|_2^2=\frac{k-1}{k},\qquad u_i^\top u_j=-\frac1k\quad(i\ne j).
\label{eq:ucalc}
\end{equation}
A coordinate-free argument shows why the cosine must take one particular value: if $\sum_cq_c^\star=0$ and every off-diagonal inner product equals $\gamma$, then $0=\|\sum_cq_c^\star\|_2^2=k+k(k-1)\gamma$, and therefore
\begin{equation}
(q_i^\star)^\top q_j^\star=-\frac1{k-1},\qquad i\ne j .
\label{eq:cos}
\end{equation}

\begin{corollary}[Canonical simplex configuration]\label{cor:etf}
The $k$ unit class directions $q_1^\star,\dots,q_k^\star$ form a regular simplex equiangular tight frame in the centered $(k-1)$-dimensional Euclidean logit space.
\end{corollary}

\begin{figure}[t]
\centering
\figbox{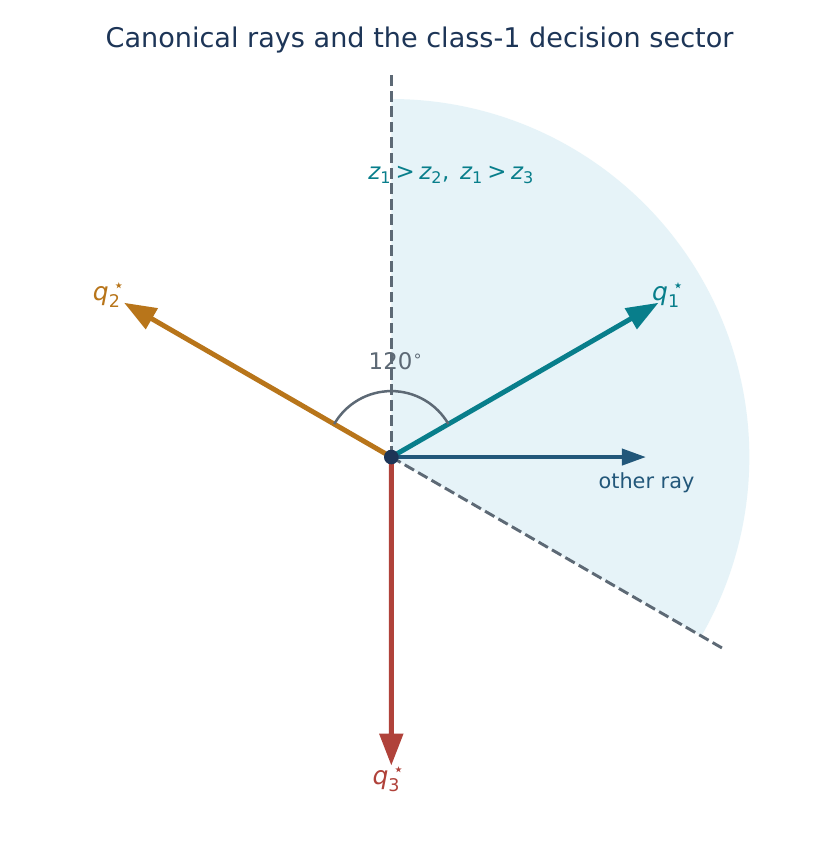}{0.6}{7cm}
\caption{Canonical rays and the class-1 decision sector in centered logits for $k=3$. Every direction inside the class-1 cone ($z_1>z_2$, $z_1>z_3$) approaches the class-1 vertex after sufficient radial scaling, but only the permutation-symmetric bisector $q_1^\star$ is the unique finite-radius CE optimum. Repeating the construction for each class gives the common equiangular configuration, with mutual angles of $120^\circ$.}
\label{fig:rays}
\end{figure}

This is a classical consequence of permutation symmetry and centering, not a new ETF theorem. The one-hot labels were already vertices of an affine simplex, and centering did not create those vertices. What the construction explains is why categorical symmetry identifies the same canonical directions in the dual-coordinate description, and why they acquire the familiar equiangular form under the chosen Euclidean logit metric (Figure~\ref{fig:rays}).

\subsection{From directions to a common-scale matrix}

The ETF statement concerns normalized directions. To obtain an equal-norm logit configuration with one representative per class we must also choose equal radii. For $\rho>0$ this yields
\begin{equation}
Z^\star(\rho)=a\,\Pi,\qquad a=\rho\sqrt{\frac{k}{k-1}},\qquad \Pi=I_k-\frac1k\one\one^\top .
\label{eq:Zstar}
\end{equation}
The rows have equal norm and the canonical angles. The scale $a$, however, is not selected by unregularized one-hot cross-entropy, since the loss keeps decreasing as the canonical radius grows. Equal radii must either be imposed as an additional assumption or follow from an explicit symmetric resource-allocation problem. As an example, prescribe a common squared-logit budget $R^2>0$ across one representative per class:
\begin{equation}
\min_{z^{(1)},\dots,z^{(k)}\in\Hsp}\ \sum_{c=1}^kL_c\big(z^{(c)}\big)\quad\text{subject to}\quad\sum_{c=1}^k\|z^{(c)}\|_2^2=R^2 .
\label{eq:budget}
\end{equation}

\begin{proposition}[Common radii under a shared budget]\label{prop:budget}
Problem \eqref{eq:budget} has the unique solution
\begin{equation}
z^{(c)\star}=\frac{R}{\sqrt k}\,q_c^\star,\qquad c=1,\dots,k,
\label{eq:budgetsol}
\end{equation}
and hence $Z^\star=a\Pi$ with $a=R/\sqrt{k-1}$.
\end{proposition}

\begin{proof}
Set $s=\|z\|_2^2$. Theorem~\ref{thm:direction} reduces each class's best achievable loss to the common function
\begin{equation}
g(s)=\log\Big(1+(k-1)e^{-\beta\sqrt s}\Big),\qquad \beta=\sqrt{\frac{k}{k-1}} .
\label{eq:g}
\end{equation}
By Appendix~A, $g$ is strictly convex on $[0,\infty)$. Jensen's inequality gives $\sum_cg(s_c)\ge k\,g(R^2/k)$, with equality only if all $s_c=R^2/k$. Theorem~\ref{thm:direction} then fixes the corresponding class directions uniquely.
\end{proof}

The result is conditional: it shows how a shared symmetric constraint can select a common radial scale, not that balancing occurs automatically in network training.

\subsection{Why a correct decision or vanishing CE does not imply an ETF}

The unconditional one-hot logit problem has infimum zero, but every finite logit has strictly positive cross-entropy:
\begin{equation}
\inf_{z\in\Hsp}L_c(z)=0,\qquad L_c(z)>0\ \text{for every finite }z .
\label{eq:inf}
\end{equation}
Different correct-class rays can approach the same target vertex. For three classes, consider
\begin{equation}
Z(t)=t\begin{pmatrix}2&0&-2\\-2&2&0\\-2&0&2\end{pmatrix},\qquad t\to\infty .
\label{eq:counter}
\end{equation}
In every row the correct-class logit dominates the other two, so each cross-entropy tends to zero. Yet the normalized row-pair cosines are $-1/2$, $-1$, and $+1/2$ for the pairs $(1,2)$, $(1,3)$, and $(2,3)$, not a common equiangular value. Vanishing classification loss therefore does not force the canonical logit ETF without further conditions. Likewise, a canonical logit configuration does not fix the geometry of the feature vectors that produce it; that is the separate question of Euclidean realization, taken up next.

\section{Euclidean Realization: From Canonical Logits to Feature Geometry}\label{sec:realization}

\subsection{The bias-free feature head}

The first two levels describe the categorical statistical model. The third begins when a neural classifier realizes logits as linear scores of a learned feature $h$. Which geometric relationships of centered logits can be attributed to features, and which depend on the head that maps one space into the other? Answering this requires studying the map itself before appealing to any ETF factorization.

For a bias-free classifier, $p=\softmax(Wh)$ with rows $w_j^\top$ of $W$. Decompose a nonzero feature as $h=rs$ with $r=\|h\|_2>0$ and $\|s\|_2=1$. The exact log-odds between the target class $c$ and a competitor $j$ are
\begin{equation}
\log\frac{p_c}{p_j}=(w_c-w_j)^\top h=r\,(w_c-w_j)^\top s .
\label{eq:featlogodds}
\end{equation}
The relevant Euclidean objects are therefore not only the classifier vectors themselves but their \emph{differences}, which define the separating hyperplanes. Substitution into the one-hot objective gives
\begin{equation}
L_c(r,s)=\log\Big(1+\sum_{j\ne c}\exp\big[-r\,(w_c-w_j)^\top s\big]\Big).
\label{eq:Lrs}
\end{equation}
When $w_c\ne w_j$, let $\theta_{cj}$ be the Euclidean angle between $s$ and $w_c-w_j$. Each exponential is then governed by the angular margin
\begin{equation}
-r\,\|w_c-w_j\|_2\cos\theta_{cj}.
\label{eq:margin}
\end{equation}
Equations \eqref{eq:featlogodds}--\eqref{eq:margin} are exact and require no equal-norm assumption on the classifier weights. They show why angular geometry matters for a bias-free classifier: class decisions are determined by the signs of directional margins, and the feature radius scales their magnitudes. If two class rows coincide, their mutual margin vanishes identically and the corresponding angle is not needed.

The strict class-$c$ decision region is the cone
\begin{equation}
C_c(W)=\{h:\ (w_c-w_j)^\top h>0\ \text{for all }j\ne c\}.
\label{eq:cone}
\end{equation}
Every positive rescaling $rs$ of an interior cone direction predicts the same class. Along a direction with all correct margins strictly positive, cross-entropy decreases to zero as $r$ grows; along a direction favoring another class, radial growth can increase the target loss. The feature-space picture thus reproduces the decision-versus-concentration distinction found in centered logits, but only after the head has transformed one direction into another.

\subsection{Which angles does the head preserve?}

\begin{figure}[t]
\centering
\figbox{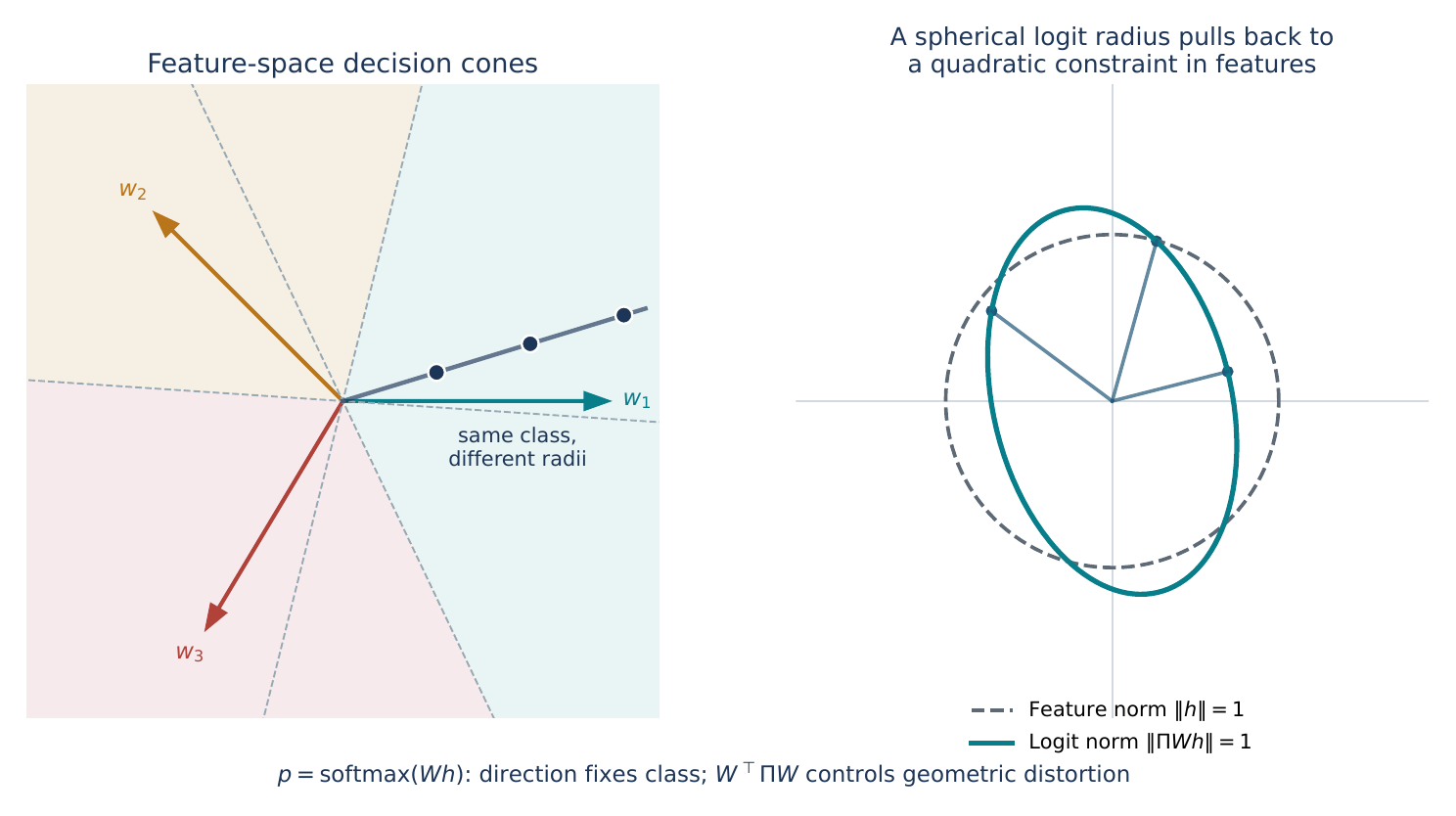}{0.9}{5cm}
\caption{Left: a bias-free head partitions feature space into conical decision regions; positive radial scaling stays inside a cone but changes the loss. Right: the Euclidean logit-radius constraint pulls back to a generally anisotropic quadratic constraint (ellipse) in feature coordinates, in contrast to the feature sphere.}
\label{fig:head}
\end{figure}

It is tempting to identify the angle of a feature with the angle of the logit vector it produces. This is false in general. After removal of the unobservable common shift, the head acts as
\begin{equation}
z=\Pi Wh=r\,\Pi Ws .
\label{eq:effhead}
\end{equation}
If $\Pi Ws\ne0$, the centered-logit radius and unit direction are
\begin{equation}
\rho=r\|\Pi Ws\|_2,\qquad q=\frac{\Pi Ws}{\|\Pi Ws\|_2}.
\label{eq:rhoq}
\end{equation}
Hence the logit radius can vary even between two features of the same Euclidean radius, because the head stretches some feature directions more than others (Figure~\ref{fig:head}). This geometry is captured exactly by the positive-semidefinite matrix
\begin{equation}
M=W^\top\Pi W .
\label{eq:M}
\end{equation}
For two features with nonzero centered logit images,
\begin{equation}
\cos\angle(\Pi Wh_1,\Pi Wh_2)=\frac{h_1^\top Mh_2}{\sqrt{h_1^\top Mh_1}\sqrt{h_2^\top Mh_2}} .
\label{eq:pullback}
\end{equation}
This is the exact pullback of Euclidean logit angles through the head. It is not the Fisher metric of the categorical simplex, and it generally differs from the original Euclidean metric on features. A fixed centered-logit radius means
\begin{equation}
\|\Pi Wh\|_2^2=\rho^2\iff h^\top Mh=\rho^2,
\label{eq:ellipsoid}
\end{equation}
which is in general an ellipsoidal or cylindrical constraint in feature coordinates, not a feature sphere. The metric for which the head is an isometry onto its image is the induced one, not the raw Euclidean metric on features.

The distinction sharpens further when the head has an invisible subspace,
\begin{equation}
\mathcal N=\ker(\Pi W).
\label{eq:kernel}
\end{equation}
If $n\in\mathcal N$, then $h$ and $h+n$ give identical categorical probabilities, so the classifier cannot distinguish features that may be far apart in Euclidean feature space. The geometry relevant to predicted probabilities lives on the quotient $\R^d/\mathcal N$, where $M$ defines a positive-definite quadratic metric. This separates the geometry the model contains from the geometry its class predictions can observe.

\subsection{Can the head realize every canonical class direction?}

The directions $u_c=e_c-b$ of Section~\ref{sec:meeting} span the centered hyperplane $\Hsp$. Whether a fixed head can produce them is a linear-algebraic question, distinct from whether the feature vectors themselves form an ETF.

\begin{proposition}[Attainability of canonical class rays]\label{prop:attain}
The class-$c$ canonical ray is attainable under $\Pi W$ if and only if $u_c\in\im(\Pi W)$. Every canonical class ray is attainable if and only if
\begin{equation}
\operatorname{rank}(\Pi W)=k-1 .
\label{eq:rank}
\end{equation}
In particular, complete attainability requires $d\ge k-1$.
\end{proposition}

\begin{proof}
Reaching the positive ray generated by $u_c$ means finding $h_c$ and $a_c>0$ with $\Pi Wh_c=a_cu_c$. Since the image of a linear map is a vector space, this is equivalent to $u_c\in\im(\Pi W)$. The class directions span $\Hsp$, so all are reachable precisely when the image equals $\Hsp$, which has dimension $k-1$.
\end{proof}

Surjectivity of $\Pi W$ guarantees access to the canonical logit rays, not uniqueness of their feature preimages: invisible directions and anisotropic stretching remain possible. The rank criterion marks the boundary between logit attainability and the much stronger question of Euclidean feature geometry.

\subsection{The factorization problem: same logits, different features}

Suppose now that one representative feature is chosen for every class and the canonical directions share a common radial scale, so that the centered logit matrix is $a\Pi$ with $a>0$. Let the rows of $H\in\R^{k\times d}$ be the class-representative features and the rows of $W\in\R^{k\times d}$ the classifier weights. The realization condition is
\begin{equation}
HW^\top=a\Pi .
\label{eq:realize}
\end{equation}
Its rank forces $d\ge k-1$, but it does not determine feature lengths, angles, or alignment. To make the ambiguity concrete, put $m=k-1$ and choose $U\in\R^{k\times m}$ with $U^\top U=I_m$ and $UU^\top=\Pi$. A particularly symmetric realization is
\begin{equation}
H_0=W_0=\sqrt a\,U,\qquad H_0W_0^\top=a\Pi .
\label{eq:sym}
\end{equation}
Its feature and classifier vectors have equal norms and ETF angles, but this representation is not forced by the product.

\begin{proposition}[Factorization ambiguity]\label{prop:ambiguity}
Suppose $H,W\in\R^{k\times d}$ both have rank $m$ and satisfy $HW^\top=a\Pi$. Then there exist full-row-rank matrices $B,C\in\R^{m\times d}$ with
\begin{equation}
H=\sqrt a\,UB,\qquad W=\sqrt a\,UC,\qquad BC^\top=I_m .
\label{eq:ambig}
\end{equation}
Conversely, any such pair has product $a\Pi$. The feature and classifier Gram matrices are
\begin{equation}
HH^\top=a\,UBB^\top U^\top,\qquad WW^\top=a\,UCC^\top U^\top,
\label{eq:gram}
\end{equation}
and need not be multiples of $\Pi$.
\end{proposition}

\begin{proof}
Since $HW^\top$ has rank $m$ and both factors have rank $m$, their column spaces equal $\im\Pi=\operatorname{col}U$. Writing them as in \eqref{eq:ambig} and multiplying the product by $U^\top$ on the left and $U$ on the right gives $BC^\top=I_m$. The converse follows by multiplication.
\end{proof}

\begin{figure}[t]
\centering
\figbox{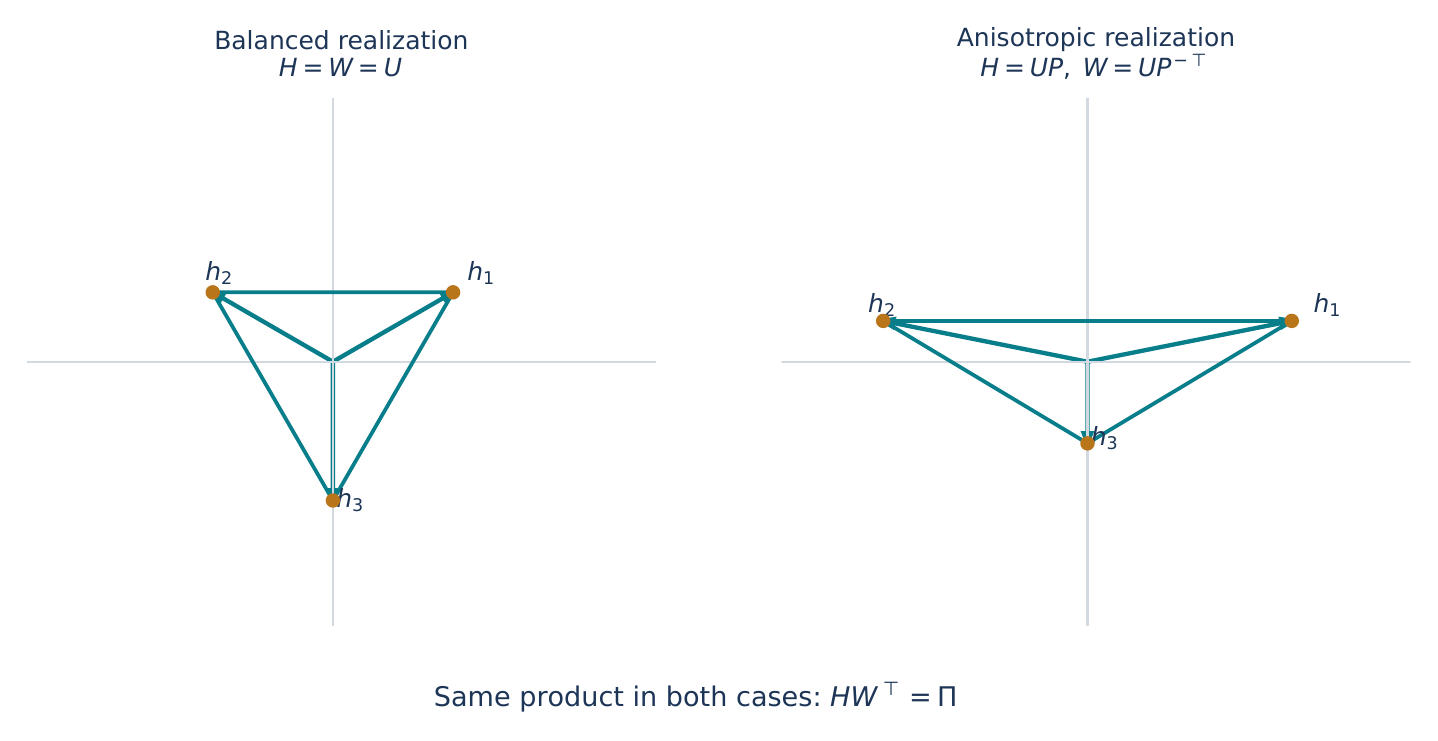}{0.9}{5cm}
\caption{Two realizations of the same canonical logit product $HW^\top=\Pi$. Left: the balanced realization $H=W=U$. Right: an anisotropic realization $H=UP$, $W=UP^{-\top}$, with $P$ invertible. The feature Gram geometry is not determined by the product alone.}
\label{fig:fact}
\end{figure}

At the minimal dimension $d=m$, any invertible $B$ together with $C=B^{-\top}$ preserves the product but in general deforms the individual feature and classifier geometries. For example, with $k=3$,
\begin{equation}
U=\begin{pmatrix}1/\sqrt2&1/\sqrt6\\-1/\sqrt2&1/\sqrt6\\0&-2/\sqrt6\end{pmatrix},\qquad B=\operatorname{diag}(t,t^{-1}),\quad C=B^{-1},\quad t>0 .
\label{eq:example}
\end{equation}
With $H=\sqrt a\,UB$, the first and third feature rows have squared norms
\begin{equation}
\|h_1\|_2^2=a\Big(\frac{t^2}{2}+\frac1{6t^2}\Big),\qquad\|h_3\|_2^2=\frac{2a}{3t^2}.
\label{eq:norms}
\end{equation}
For generic $t$ these differ, although the logits remain exactly $a\Pi$. A network can thus realize the canonical logit ETF while its feature vectors are not an ETF (Figure~\ref{fig:fact}).

\subsection{Balance as an additional realization principle}

How, then, can canonical logit geometry become an aligned feature and classifier ETF? The factorization problem needs an extra selection rule. A standard one is to minimize the combined squared Frobenius norm of the two factors; the following is the classical balanced-factorization identity specialized to our product.

\begin{theorem}[Minimum-norm factorization]\label{thm:minnorm}
For $d\ge k-1$ and $a>0$,
\begin{equation}
\min_{HW^\top=a\Pi}\ \frac12\big(\|H\|_F^2+\|W\|_F^2\big)=a(k-1),
\label{eq:minnorm}
\end{equation}
with equality precisely when
\begin{equation}
H=W=\sqrt a\,UR,\qquad R\in\R^{(k-1)\times d},\quad RR^\top=I_{k-1}.
\label{eq:minsol}
\end{equation}
\end{theorem}

\begin{proof}
The product has fixed trace $\tr(HW^\top)=a(k-1)$. Therefore
\begin{equation}
\frac12\big(\|H\|_F^2+\|W\|_F^2\big)=a(k-1)+\frac12\|H-W\|_F^2 .
\label{eq:identity}
\end{equation}
Equality in the lower bound requires $H=W$, and then $HH^\top=a\Pi$ yields the stated parametrization. Conversely, each such factorization attains the bound.
\end{proof}

The ETF rows arise from $HH^\top=WW^\top=a\Pi$, and the classwise alignment from $H=W$. The theorem does not predict that weight decay in an arbitrary nonlinear network imposes the canonical product or produces balanced factors; it identifies a realization principle that would suffice once the product is specified. Appendix~B records the corresponding balance statement for minimal-rank realizations.

\subsection{Where the construction meets neural-collapse geometry}

Neural collapse describes properties observed in trained representations. The present account reaches a matching classwise Euclidean configuration through a sequence of conditional geometric statements, and the logical order of these statements must not be reversed. Table~\ref{tab:status} separates what is established at each level from what is not implied.

\begin{table}[t]
\centering\small
\caption{Logical status of the geometric conclusions.}
\label{tab:status}
\begin{tabular}{@{}p{0.22\textwidth}p{0.36\textwidth}p{0.34\textwidth}@{}}
\toprule
Geometric conclusion & What is established & What is not implied\\
\midrule
Target/prediction geometry & Label vertices and KL comparison & Any prescribed feature configuration\\
Canonical logit directions & Symmetry-fixed ray; unique fixed-radius CE optimum & Every near-zero-loss sequence follows the ray\\
Equal-scale canonical logits & $Z=a\Pi$ under equal radii or a shared budget & Unregularized CE chooses a finite common radius\\
Availability under a fixed head & All canonical rays iff $\operatorname{rank}(\Pi W)=k-1$ & Unique feature preimages or angle preservation\\
Feature ETF (NC2) & A balanced factorization realizes it & Arbitrary factors of the same logits are ETFs\\
Feature--classifier alignment (NC3) & Holds under Theorem~\ref{thm:minnorm} & Logits alone determine alignment\\
Within-class collapse (NC1) & Not established here & The training model forces classwise feature collapse\\
Nearest-center rule (NC4) & Holds for aligned, equal-norm class prototypes & Actual network training converges to them\\
\bottomrule
\end{tabular}
\end{table}

Indeed, if $w_c=\gamma h_c$ with $\gamma>0$ and all class prototypes $h_c$ have the same norm, then
\begin{equation}
\argmax_c\ w_c^\top h=\argmin_c\ \|h-h_c\|_2^2 .
\label{eq:nearest}
\end{equation}
The framework thus meets neural-collapse geometry at a distinguished Euclidean realization of canonical classification logits. It does not assert that all training problems reach that realization, that any finite-loss model exhibits neural collapse, or that the within-class configuration has been explained.

\subsection{Remark on affine classifiers}

An affine head $z=Wh+\beta$ becomes a linear map after augmenting the feature vector:
\begin{equation}
\tilde h=\begin{pmatrix}h\\1\end{pmatrix},\qquad\tilde W=\begin{pmatrix}W&\beta\end{pmatrix},\qquad z=\tilde W\tilde h .
\label{eq:affine}
\end{equation}
This is an algebraic equivalence, not a preservation of the original angular--radial feature geometry: the augmented features lie in the affine hyperplane with last coordinate $1$, and $h\mapsto rh$ does not correspond to scaling $\tilde h$ by $r$. Conical decision regions in the original feature space are therefore a property of the bias-free head, not an automatic consequence of homogeneous-coordinate notation. (The bias is written $\beta$ here to avoid confusion with the barycenter $b$.)

\section{Discussion: What Has Been Geometrically Constructed?}\label{sec:discussion}

\subsection{Specification, selection, and realization}

The construction crosses two different kinds of mathematical link. Between probability distributions and centered logits, Legendre duality provides alternative coordinates for the same interior statistical model and the same KL-based objective. Between logits and features, a model-dependent linear map generates scores from representations and may distort or erase Euclidean directions. These two connections should not be conflated.

Within the first connection, permutation symmetry singles out the canonical class rays independently of the objective, and cross-entropy then assigns those rays a conditional optimality at every prescribed Euclidean logit radius. Across classes, the orbit of rays is equiangular. Neither an arbitrary near-zero-loss sequence nor an arbitrary factorization of $a\Pi$ need exhibit this geometry. These boundaries of implication are part of the geometric explanation itself.

\subsection{Angular--radial geometry as an interpretive instrument}

Angles are primary for categorical decisions in centered logit space: scaling a fixed nonzero logit direction does not change the ordering of the classes. They are equally central in bias-free feature space, where classifier-difference vectors define conical regions. Radial scale, however, cannot be removed from the probabilistic objective: it changes log-odds gaps and entropy, and may increase correct or incorrect confidence depending on direction. The induced form $M=W^\top\Pi W$ explains why an angular relation in logits need not coincide with the Euclidean angle between features.

Angular--radial separation is accordingly proposed neither as a replacement for information geometry nor as an independent source of the ETF. It is a lens that shows what classification uses direction for, what magnitude contributes, and where those roles are transformed in passing between geometric levels.

\subsection{Limitations and directions}

The construction is confined to standard one-hot softmax cross-entropy, a chosen constant Euclidean metric on centered logits, and a bias-free final head. The fixed-radius result optimizes free class logits, whereas an actual head may realize only part of the centered-logit hyperplane, and the factorization statements concern a particular target product under additional norm or balance assumptions. We do not establish a training-convergence theorem, within-class collapse, or a temporal phase transition.

The framework nevertheless suggests precise follow-up questions: which structures change for other class encodings or convex potentials; how anisotropic logit metrics alter the class directions; which geometries persist under class imbalance; and which feature metrics permit faithful realization of canonical logit angles. Each requires its own assumptions. Appendix~E shows, for an interior target, how the radial but not the directional configuration changes.

\section{Conclusion}\label{sec:conclusion}

Multiclass classification has a geometric structure before any training trajectory is considered. Labels and predictions define a categorical probability problem; negative entropy and log-sum-exp equip its probability and natural-logit descriptions with a common convex-dual organization; and the one-hot objective is a directed KL divergence that can equally be read as a Fenchel--Young gap. Angular--radial decomposition then shows how the dual coordinates separate categorical ordering from the scale of log-odds.

The loss-independent geometric path follows the symmetry of the labels from the uniform distribution toward a target vertex and obtains a canonical centered-logit ray. The cross-entropy path proves that the same positive ray uniquely minimizes the classification divergence at every fixed Euclidean logit radius. Their agreement reflects a common class symmetry, whose classwise orbit yields the regular simplex ETF once Euclidean angles are introduced. Equal-scale logit configurations and aligned feature and classifier ETFs require additional, separate conditions.

Studying the head $\Pi W$ directly identifies the feature directions visible to classification, the metric induced by the logits, the criterion for attaining the canonical rays, and the ambiguity of their Euclidean factorizations. Balanced minimum-norm realization provides a distinguished meeting point with neural-collapse geometry. The result is a progressive construction from the geometry specified by classification toward the geometry a model can realize, with explicit mathematical boundaries between the two.

\appendix
\section{Strict Convexity of the Shared-Budget Radial Function}

Let $b_0=k-1>0$, $\beta=\sqrt{k/(k-1)}>0$, and
\begin{equation}
g(s)=\log\big(1+b_0e^{-\beta\sqrt s}\big),\qquad s\ge0 .
\label{eq:A1}
\end{equation}
Set $\phi(t)=\log(1+b_0e^{-\beta t})$ for $t\ge0$. Then
\begin{equation}
\phi'(t)=-\beta\frac{b_0e^{-\beta t}}{1+b_0e^{-\beta t}}<0,\qquad
\phi''(t)=\beta^2\frac{b_0e^{-\beta t}}{(1+b_0e^{-\beta t})^2}>0 .
\label{eq:A2}
\end{equation}
For $s>0$,
\begin{equation}
g''(s)=\frac{\phi''(\sqrt s)}{4s}-\frac{\phi'(\sqrt s)}{4s^{3/2}}>0 .
\label{eq:A3}
\end{equation}
The continuous extension at $s=0$ remains strictly convex on $[0,\infty)$: $\phi$ is strictly decreasing and strictly convex while $s\mapsto\sqrt s$ is strictly concave, so for two distinct values of $s$ and $0<\lambda<1$, strict concavity followed by strict decrease and convexity gives the strict convexity inequality. Hence $\sum_cg(s_c)$ is uniquely minimized at equal allocation under $\sum_cs_c=R^2$.

\section{A Balance Fact for Minimal-Rank Realizations}

Assume the setting of Proposition~\ref{prop:ambiguity} and let $H=\sqrt a\,UA$, $W=\sqrt a\,UB$ with $A,B$ of full row rank and $AB^\top=I_r$. If $H^\top H=W^\top W$, then $A^\top A=B^\top B$. Since both have full row rank, there exists an orthogonal $Q\in O(r)$ with $B=QA$. The product condition gives
\begin{equation}
AA^\top Q^\top=I_r\ \Longrightarrow\ AA^\top=Q .
\label{eq:B1}
\end{equation}
The matrix $AA^\top$ is symmetric positive definite and $Q$ is orthogonal; therefore $Q=I_r$, $AA^\top=I_r$, and $B=A$. Balanced minimal-rank realizations thus satisfy $H=W$ and are precisely the ETF factors of Theorem~\ref{thm:minnorm}. This is an algebraic statement, not a prediction that arbitrary network optimization establishes balance.

\section{Pullback Geometry and Class-Ray Realizability}

For arbitrary feature vectors $h_1,h_2$, the centered logit inner product is
\begin{equation}
\langle\Pi Wh_1,\Pi Wh_2\rangle=h_1^\top W^\top\Pi^\top\Pi Wh_2=h_1^\top Mh_2,\qquad M=W^\top\Pi W,
\label{eq:C1}
\end{equation}
because $\Pi^\top\Pi=\Pi$. The null space of $M$ equals $\ker(\Pi W)$, since $h^\top Mh=\|\Pi Wh\|_2^2$, and the induced inner product is strictly positive definite on the quotient by this null space. For $h=rs$ the transformed logit radius is $\rho=r\sqrt{s^\top Ms}$; it depends on direction unless $M$ restricts to a scalar multiple of the identity on the subspace under study. If $\operatorname{rank}(\Pi W)<k-1$, the image cannot contain all $k$ canonical class directions simultaneously, because they span $\Hsp$.

\section{Three-Class Verification and a Noncanonical Ray}

For $k=3$ the centered directions are
\begin{equation}
u_1=\tfrac13(2,-1,-1),\quad u_2=\tfrac13(-1,2,-1),\quad u_3=\tfrac13(-1,-1,2).
\label{eq:D1}
\end{equation}
Each has squared norm $2/3$ and every distinct pair has inner product $-1/3$, giving cosine $-1/2$. For target class $1$ and fixed radius $\rho$,
\begin{equation}
z_1^\star=\frac{\rho}{\sqrt6}(2,-1,-1),\qquad \softmax(z_1^\star)_1=\frac1{1+2e^{-\sqrt{3/2}\,\rho}} .
\label{eq:D2}
\end{equation}
The alternative centered ray $z=t(2,0,-2)$ also approaches the same target vertex as $t\to\infty$, but it lacks off-target permutation symmetry and is not the CE optimum at fixed Euclidean radius.

\section{Interior Symmetric Targets and Finite Canonical Scale}

One-hot supervision requires a boundary prediction and therefore has no finite centered-logit optimum. To see how an interior target changes this without changing the canonical symmetric direction, take
\begin{equation}
v_c^{(\varepsilon)}=(1-\varepsilon)\,e_c+\frac\varepsilon k\one,\qquad0<\varepsilon<1 .
\label{eq:E1}
\end{equation}
This distribution lies on the class-symmetric barycenter-to-vertex segment at $t=1-\varepsilon$. Since its coordinates are strictly positive, its KL optimum is attained at the finite centered inverse-softmax coordinate
\begin{equation}
z_c^\star=\log\Big(\frac{k-(k-1)\varepsilon}{\varepsilon}\Big)\Big(e_c-\frac1k\one\Big)=\Pi\log v_c^{(\varepsilon)} .
\label{eq:E2}
\end{equation}
The canonical direction is unchanged, whereas the radial coefficient is finite and depends nonlinearly on the smoothing parameter. The directly centered target is $(1-\varepsilon)u_c$, which should not be confused with the dual logit coefficient in \eqref{eq:E2}. This is an elementary inverse-softmax calculation and not a claim about feature-space neural collapse under label smoothing.


\end{document}